%% file: _main.tex
\input{_constants}
\arxiv 

\pdfoutput=1
\documentclass[10pt,twocolumn,letterpaper]{article}
\input{cvpr_header}
\begin{document}
\title{A Unified Diffusion Framework for Scene-aware Human Motion Estimation from Sparse Signals}
\author{\authorBlock}
\maketitle

\input{paper_v1}

{\small
\bibliographystyle{ieeenat_fullname}
\bibliography{11_references}
}

\ifarxiv \clearpage \appendix \input{12_appendix} \fi

\end{document}

%% file: _constants.tex
\def\paperID{6923}
\def\confName{CVPR}
\def\confYear{2024}

\def\authorBlock{
    Jiangnan Tang$^1$\qquad
    Jingya Wang$^{1,2,}$\footnotemark[1] \qquad
    Kaiyang Ji$^1$ \qquad
    Lan Xu$^{1,2}$ \qquad
    Jingyi Yu$^{1,2}$ \qquad
    Ye Shi$^{1,2,}$\thanks{Corresponding author} \\
    $^1$ShanghaiTech University \\ $^2$Shanghai Engineering Research Center of Intelligent Vision and Imaging \\
    {\tt\small \{tangjn2022, wangjingya, jiky, xulan1, yujingyi, shiye\}@shanghaitech.edu.cn}
}

\newif\ifreview 
\newif\ifarxiv \newcommand{\arxiv}{\arxivtrue}
\newif\ifcamera 
\newif\ifrebuttal 

%% file: cvpr_header.tex
\ifreview \usepackage[review]{cvpr} \fi
\ifarxiv \usepackage[pagenumbers]{cvpr} \fi
\ifrebuttal \usepackage[rebuttal]{cvpr} \fi
\ifcamera \usepackage{cvpr} \fi

\input{_macros}  

\usepackage{xr-hyper}

\makeatletter
\newcommand*{\addFileDependency}[1]{
  \typeout{(#1)}
  \@addtofilelist{#1}
  \IfFileExists{#1}{}{\typeout{No file #1.}}
}

\makeatother

\definecolor{cvprblue}{rgb}{0.21,0.49,0.74}
\usepackage[pagebackref,breaklinks,colorlinks,citecolor=cvprblue]{hyperref}
\usepackage[capitalize]{cleveref}
\usepackage{xcolor}
\usepackage[linesnumbered,ruled,vlined]{algorithm2e}
\usepackage{setspace}
\usepackage[accsupp]{axessibility}
\crefname{section}{Sec.}{Secs.}
\crefname{table}{Table}{Tables}
\crefname{figure}{Fig.}{Figs.}
\crefname{algorithm}{Alg.}{Algos.}

\SetCommentSty{mycommfont}

\SetKwInput{KwInput}{Input}                
\SetKwInput{KwOutput}{Output}              

%% file: _macros.tex
\usepackage{graphicx}	
\usepackage{amsmath}	
\usepackage{amssymb}	
\usepackage{booktabs}
\usepackage{times}
\usepackage{microtype}
\usepackage{epsfig}
\usepackage[table,xcdraw,dvipsnames]{xcolor}
\usepackage{caption}
\usepackage{float}
\usepackage{placeins}
\usepackage{color, colortbl}
\usepackage{stfloats}
\usepackage{enumitem}
\usepackage{tabularx}
\usepackage{xstring}
\usepackage{multirow}
\usepackage{xspace}
\usepackage{url}
\usepackage{subcaption}
\usepackage{xcolor}
\usepackage[hang,flushmargin]{footmisc}

\ifcamera \usepackage[accsupp]{axessibility} \fi

% ...

\ifarxiv  \fi

\newcommand{\R}[1]{{%
    \textbf{%
        \ifstrequal{#1}{1}{\textcolor{red}{R#1}}{%
        \ifstrequal{#1}{2}{\textcolor{blue}{R#1}}{%
        \ifstrequal{#1}{3}{\textcolor{magenta}{R#1}}{%
        \ifstrequal{#1}{4}{\textcolor{teal}{R#1}}{%
                           \textcolor{cyan}{R#1}%
        }}}}%
    }%
}}

%% file: paper_v1.tex
\begin{abstract}
Estimating full-body human motion via sparse tracking signals from head-mounted displays and hand controllers in 3D scenes is crucial to applications in AR/VR. One of the biggest challenges to this task is the one-to-many mapping from sparse observations to dense full-body motions, which endowed inherent ambiguities. To help resolve this ambiguous problem, we introduce a new framework to combine rich contextual information provided by scenes to benefit full-body motion tracking from sparse observations. To estimate plausible human motions given sparse tracking signals and 3D scenes, we develop $\text{S}^2$Fusion, a unified framework fusing \underline{S}cene and sparse \underline{S}ignals with a conditional dif\underline{Fusion} model. 
$\text{S}^2$Fusion first extracts the spatial-temporal relations residing in the sparse signals via a periodic autoencoder, and then produces time-alignment feature embedding as additional inputs. Subsequently, by drawing initial noisy motion from a pre-trained prior, $\text{S}^2$Fusion utilizes conditional diffusion to fuse scene geometry and sparse tracking signals to generate full-body scene-aware motions. The sampling procedure of $\text{S}^2$Fusion is further guided by a specially designed scene-penetration loss and phase-matching loss, which effectively regularizes the motion of the lower body even in the absence of any tracking signals, making the generated motion much more plausible and coherent. Extensive experimental results have demonstrated that our $\text{S}^2$Fusion outperforms the state-of-the-art in terms of estimation quality and smoothness. Code is available at \url{https://github.com/jn-tang/S2Fusion}.
\end{abstract}

\section{Introduction}
\label{sec:intro}

With the emergence of advanced AR/VR technologies, there is an increasing demand for generating realistic human avatars in applications such as virtual conferencing and gaming. However, common AR/VR devices, e.g., HTC Vive and Meta Quest Pro, provide only sparse tracking signals from inertial measurement units (IMU) embedded in single head-mounted displays (HMD) and hand controllers. Using sparse tracking signals to generate dense full-body motions is a one-to-many mapping with inherent ambiguities, making this problem a challenging task. 

A natural method to generate full-body motion is leveraging data-driven methods that utilize both large-scale motion capture data \cite{mahmood2019amass} and sparse tracking signals by AR/VR devices. Recently, various data-driven models -- ranging from simple regression-based methods \cite{jiang2022avatarposer, yang2021lobstr, di2023dual, zheng2023realistic, aliakbarian2023hmd} to probabilistic generative models such as VAE \cite{Dittadi_2021_ICCV}, normalizing flow \cite{aliakbarian2022flag} and diffusion model \cite{du2023agrol} -- are deployed. Although these methods show various ways of finding the most probable human motion estimation, they still fail to narrow down
the distribution of possible motion space, leaving the crux of one-to-many ambiguity unresolved. Existing methods \cite{aliakbarian2022flag, di2023dual, Dittadi_2021_ICCV, du2023agrol, jiang2022avatarposer, winkler2022questsim, ye2022neural3points} also failed to pose constraints on the pose of the lower body, making the generated motion disastrous as the legs are free to penetrate through scene geometries; the generated motion of legs may also uncorrelated with hands motion, resulting in the moving of hands and feet on the same side. The incurring of implausible leg motions is due to the lack of observations on the lower body. 

\begin{figure}
    \centering
    \resizebox{\columnwidth}{!}{\includegraphics{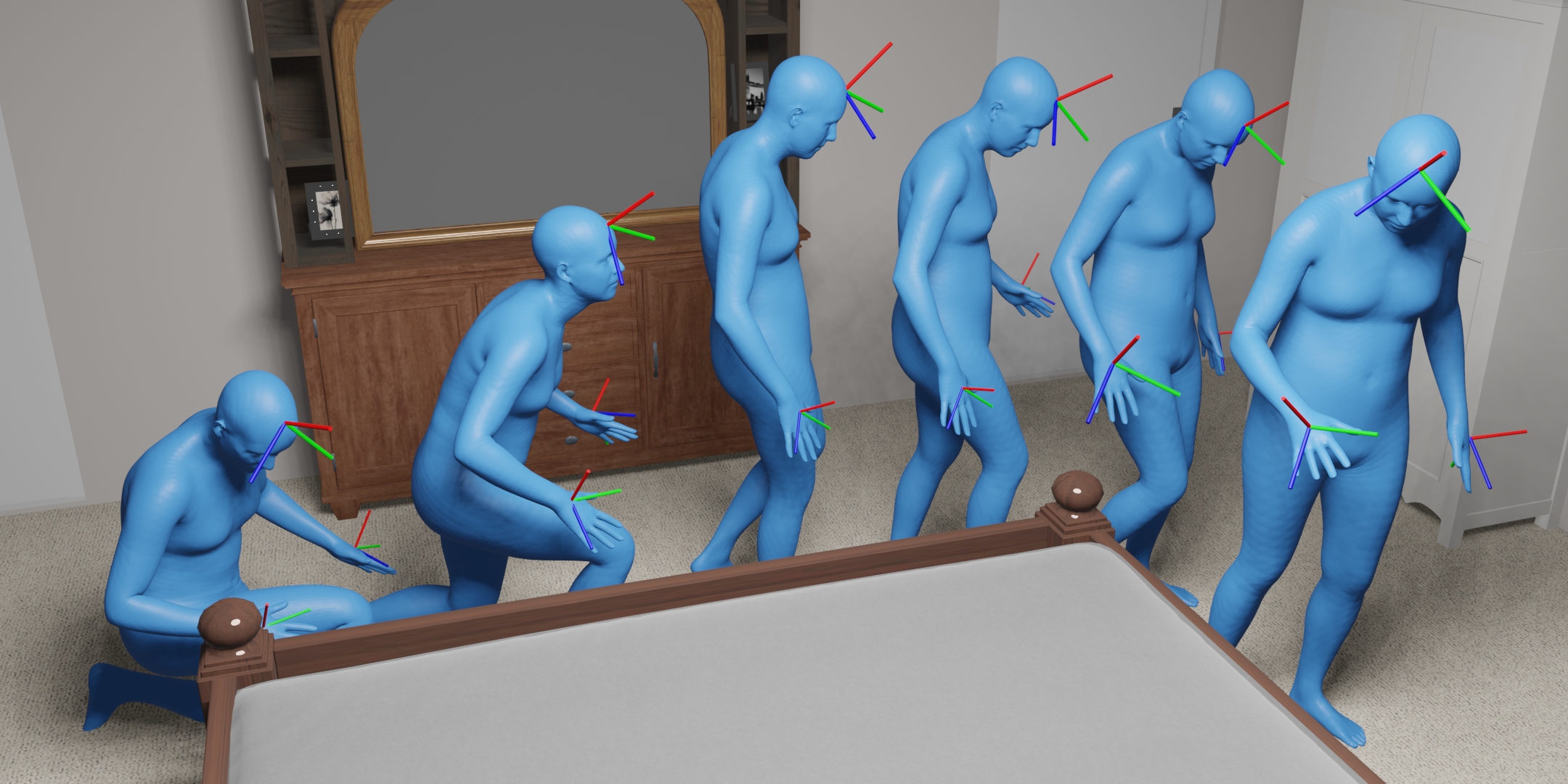}}
    \caption{Given sparse tracking signals from only the head and left/right hands, our method accurately estimates full-body motion in the 3D scene.}
    \label{fig:main}
\end{figure}

Since human motion is highly related to the surrounding environments, introducing the scene modality can greatly reduce the ambiguities in estimating full-body motion from sparse tracking signals. The rich contextual information provides valuable cues to infer the lower body motion, even though no direct observations are available. Therefore, we propose to combine the scene information with sparse tracking signals for human motion estimation. To handle the uncorrelated issues between the generated motion of legs and hands, we use the temporal movement pattern presented in the upper body observations to control the generation of lower body motions.


To estimate plausible human motions given sparse tracking signals and 3D scenes, we develop $\text{S}^2$Fusion, a unified framework fusing \underline{S}cene and sparse \underline{S}ignals with a conditional dif\underline{Fusion} model. $\text{S}^2$Fusion models this specific task by a conditional diffusion model, in light of the recent success of diffusion models in motion generation \cite{tevet2022human, chen2023executing}. Diffusion models support flexible control on generating samples, by incorporating various loss functions in the loss-guided sampling process \cite{song2023loss, chung2023diffusion}. To tackle the lack of paired motion-scene datasets and to generate more diverse motion, the reverse diffusion process starts from a non-Gaussian motion distribution, by adopting a pre-trained motion prior on a large-scale motion dataset \cite{mahmood2019amass}. To coordinate the human motion in 3D scenes, we extract the periodic motion feature of the sparse tracking signals using a periodic autoencoder \cite{starke2022deepphase}; the periodic motion feature represents the alignment of full-body motions in time and space, resulting in more effective positional embeddings. To facilitate the mitigating of unrealistic lower body motions, we guide the diffusion sampling process by the gradient of our specially designed loss functions, under the framework of loss-guided sampling \cite{song2023loss}. Our loss functions include scene-penetration loss and phase-matching loss, which regularizes the lower body motion \textit{even in the absence of any tracking signals}.

In summary, our paper makes the following contributions:
\begin{itemize}[leftmargin=.25in]
    \item We introduce a new framework that combines scene information and sparse tracking signals for human motion estimation to greatly reduce the inherent ambiguities in estimating full-body motion. To further enhance human motion estimation, we propose to extract periodic motion features from sparse tracking signals to improve the coordination between the upper and lower body. 
    \item We develop a unified diffusion method, i.e.,  $\text{S}^2$Fusion, tailored for the scene-aware human motion estimation with sparse signals. $\text{S}^2$Fusion integrates three main components: 1) a motion prior that provides initial value for the reverse diffusion process, which significantly improves generation quality and inference speed; 2) a periodic motion feature extractor that learns the spatial-temporal alignment of input signals varies in unit and scale; the extracted feature is then combined with sparse tracking signals and scene geometry as conditional inputs, fed to motion generation process; 
    3) a set of specially designed loss functions, which effectively introduce regularization on the motion of the lower body during the loss-guided diffusion sampling process. 
    \item We conduct extensive experiments for the scene-aware human motion estimation and verify $\text{S}^2$Fusion can successfully reduce the inherent ambiguities posed by the sparse-to-dense problem in contrast to other pure sensor-based methods and achieve a notable performance boost. Moreover, comprehensive ablation studies also demonstrate the effectiveness of the developed three main components of $\text{S}^2$Fusion.  
\end{itemize}

\section{Related Work}
\label{sec:related}

\subsection{Motion Estimation from Sparse Sensors}

There is a surging interest in studying reconstructing human motion from sparse sensor inputs, as it is not only economical compared to traditional marker-based optical solutions or vision-based solutions \cite{zhang2023ikol, zhang2022mutual}, but also suits the needs of AR/VR applications. A major problem in reconstructing full-body motions where only sparse observations are available is the ambiguities inherent in this one-to-many mapping problem. One line of work reconstructs full-body motion from six IMUs located on the head, left/right wrists, left/right knees, and torso, pioneered by von Marcard et al., \cite{von2017sparse}. Yi et al. proposed a real-time RNN-based method TransPose \cite{TransPoseSIGGRAPH2021} to predict full-body motions, and their follow-up works \cite{PIPCVPR2022, EgoLocate2023} refines the predicted motions by physics-based optimization and egocentric SLAM system, respectively. Other backbones are also explored, DIP \cite{huang2018deep} is powered by biRNN, and TIP \cite{jiang2022transformer} deployed Transformer encoder architecture. Another line of work reconstructs full-body motion from three 6D trackers located on the head and left/right hands, providing accurate translation and rotation measurements. \cite{aliakbarian2022flag, Dittadi_2021_ICCV, du2023agrol} tried various generative probabilistic models to solve the unconstrained problem, while \cite{jiang2022avatarposer, zheng2023realistic, aliakbarian2023hmd} used Transformer backbone to regress full-body motion directly. Our method fits this line of research of estimating full-body motion from three 6D trackers. In addition, our approach brings scene information to resolve the inherent ambiguities induced by sparse-to-dense mapping, resulting in notable estimation improvement. 

\subsection{Diffusion-based Probabilistic Models}


Diffusion-based probabilistic models (DPMs) \cite{ho2020denoising, song2020score} have recently shown promising results in image generation \cite{dhariwal2021diffusion}, video generation \cite{ho2022imagen, esser2023structure}, and speech synthesis \cite{kong2020diffwave}, demonstrating their powerful probabilistic modeling capabilities. The success of DPMs is mainly attributed to their ability to support versatile conditioning and are highly controllable. Increasing interest grows to further control the generation process of DPMs by adjusting the denoising trajectory, such as classifier guidance \cite{dhariwal2021diffusion}, imputation and inpainting \cite{chung2022improving, chung2023diffusion, zhu2023denoising, song2021solving}, which can be collectively termed as \textit{loss-guided diffusion} (LGD) \cite{song2023loss}. The loss-guided diffusion enables flexible conditioning without retraining existing models, effectively rendering DPMs a powerful generation tool. Our $\text{S}^2$Fusion is a specifically designed diffusion model for scene-aware human motion estimation from
sparse signals. It differentiates from standard diffusion models mainly in three components: 1) a motion prior module for diffusion initialization; 2) a comprehensive condition acquisition module; and 3) a set of specifically designed loss functions in the diffusion sampling stage. 

\subsection{Human Motion Generation}
Human motion can be generated by any signal that describes the motion, including text \cite{balaji2022ediffi, guo2022generating, zhang2022motiondiffuse, tevet2022human, chen2023executing, yuan2023physdiff, karunratanakul2023guided, shafir2023human, wu2024thor, tian2024gaze}, audio \cite{li2021ai, tseng2023edge}, action labels \cite{guo2020action2motion, petrovich2021action, tevet2022human, chen2023executing}, or unconditioned \cite{zhao2020bayesian, zhang2020perpetual, shafir2023human}. With increasingly available scene-motion datasets \cite{hassan2019resolving, taheri2020grab, bhatnagar2022behave, huang2022capturing, zhang2023neuraldome}, there are shifts in trends to generate human motions in 3D scenes, ranging from holistically generating human motion with scene-awareness \cite{wang2021synthesizing, weng2021holistic, zhang2021learning, wang2022towards, huang2023diffusion, huo2023stackflow, cong2023weakly} to fine-grained reconstructing accurate human-object interactions \cite{karunratanakul2020grasping, cao2021reconstructing, tripathi20233d, kulkarni2023nifty, xu2023interdiff, liang2024intergen, diller2023cg, peng2023hoi}. Our method is the first attempt to fuse scene information with sparse tracking signals to estimate full-body physical-plausible human motion with a sophisticated diffusion model. 

\subsection{Motion Frequency Analysis}
The characteristic of motion in the frequency domain has been utilized for motion synthesis \cite{liu1994hierarchical}, editing \cite{bruderlin1995motion, kenwright2015quaternion}, stylization \cite{unuma1995fourier}, and compression \cite{beaudoin2007adapting}. The features in the frequency domain present a holistic view of the underlying motion and are consistent over a period of time, hence are a great medium for summarizing the movement pattern. Starke et al. \cite{starke2022deepphase} have recently achieved success in using a deep neural network to learn the periodic feature from motion datasets and synthesizing high-quality motions, they demonstrate the power of a new network architecture called periodic autoencoder (PAE) in extracting the periodic behavior of the motions. Shi et al. \cite{Shi_2023_ICCV} built upon PAE a conditional VAE, which also leverages phase features in robustly generating motions. Instead of only leveraging the phase feature residing on the frequency domain, our method explores using both the temporal and frequency domain features; the temporal domain features are useful in aligning motions time axis, and the frequency domain features enabled us to correlate the upper and lower body motions.

\section{Human Motion Estimation with $\text{S}^2$Fusion}
\label{sec:method}
\label{subsec:formulation}

\label{subsec:S2f} 

\begin{figure*}
    \centering
    \resizebox{2\columnwidth}{!}{\includegraphics{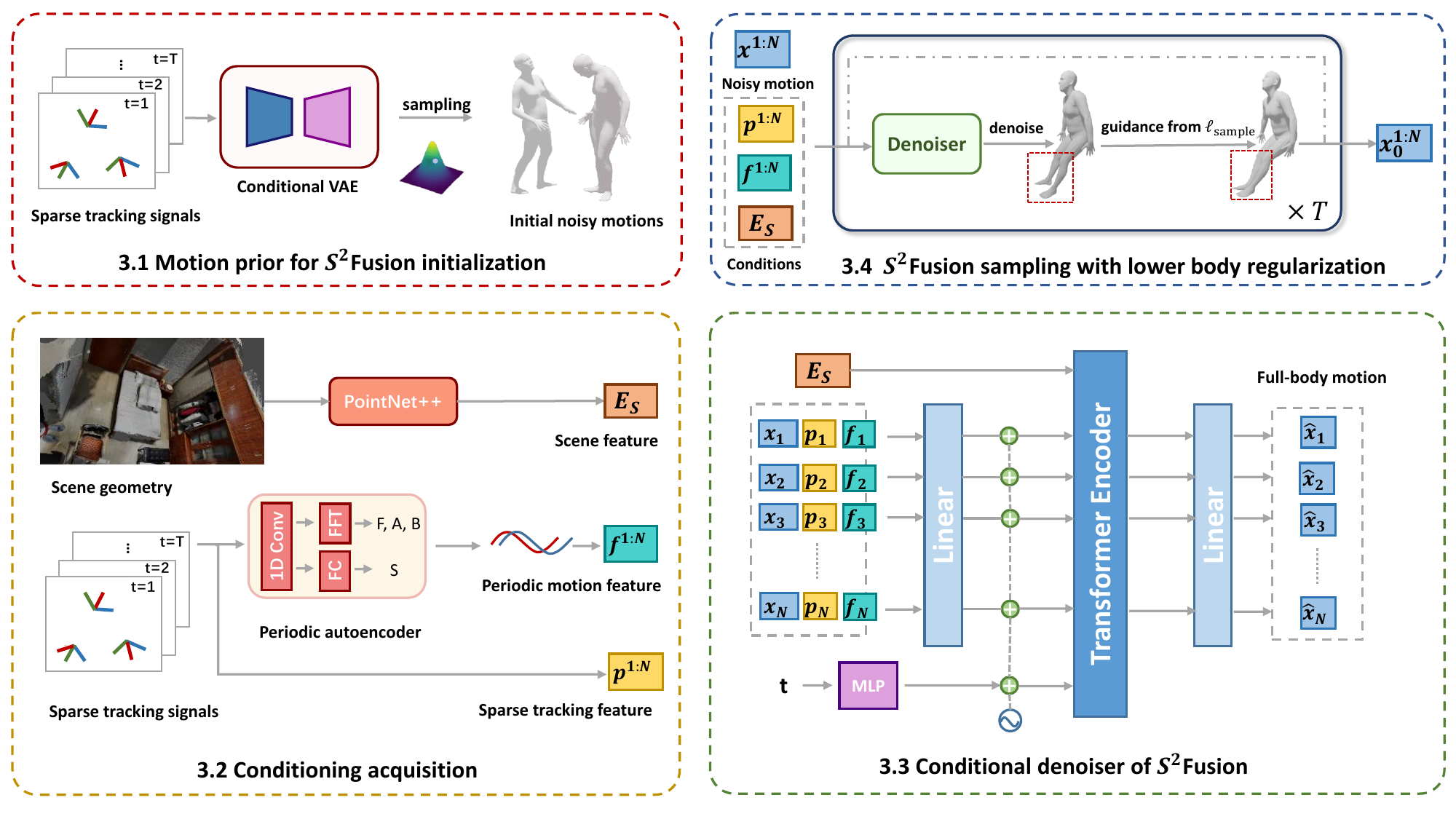}}
    \caption{Illustration of $\text{S}^2$Fusion pipeline. Given the sparse tracking signals $\mathbf{p}^{1:N}$ and scene geometry $\mathcal{S}$, $\text{S}^2$Fusion generates full-body motion with scene awareness and coherent upper and lower body movements. (1) The pre-trained motion prior $f_\phi$ first samples the initial noisy motion $\tilde{\mathbf{x}}^{1:N}$ for the reverse diffusion process; (2) then the periodic motion features $\mathbf{f}^{1:N}$ are extracted by a periodic autoencoder, and combined with encoded scene feature $\mathbf{E}_{\mathcal{S}}$ and the sparse tracking signals $\mathbf{p}^{1:N}$ to form the final conditioning input $\mathbf{c}$ to the reverse diffusion process; (3) the conditional diffusion model predicts the clean motion $\mathbf{x}^{1:N}_0$ from noisy motion $\tilde{\mathbf{x}}^{1:N}$ conditioned on $\mathbf{c}$; (4) the diffusion sampling process is further guided by the gradient of $\ell_{\text{penetration}}$ and $\ell_{\text{phase}}$ to generate scene-aware and physically plausible motions.}
    \label{fig:pipeline}
\end{figure*}

It is not trivial to estimate full-body motion from scene geometry and sparse signals, as it is an ill-conditioned one-to-many problem, and care must be taken to deal with \textit{spatial} scene geometry and \textit{temporal} sparse signals. In light of the recent success of the diffusion model in text-to-motion generation \cite{tevet2022human}, we design a conditional diffusion model to generate full-body poses given scene geometry and sparse tracking signals. However, we observed several issues that hinder the motion generation quality from directly applying the conditional diffusion model. First, due to the limited data volume of the existing motion-scene datasets, the generated motion is less diverse and unrealistic. Second, the tracking signals only come from the upper body, so it is hard to generate correlated upper and lower body motions. To tackle these challenges, we designed a novel diffusion method, i.e., $\text{S}^2$Fusion that leverages a pre-trained motion prior and augments scene and sparse tracking signals with periodic alignment features to generate more diverse and realistic motions. 

Given a sequence of sparse tracking signals $\mathbf{p}^{1:N} \in \mathbb{R}^{N \times c}$ and scene geometry $\mathbf{S} \in \mathbb{R}^{P \times 3}$, we aim to predict full-body motion $\mathbf{x}^{1:N} \in \mathbb{R}^{N \times n}$. Here $c$ and $n$ respectively denote the dimension of input and output, and $P$ denotes the number of points in the scene point cloud. The popular SMPL \cite{loper2023smpl} model is used to represent human poses and only the first 22 joints of the SMPL model \cite{jiang2022avatarposer} are considered. We use the 6D representation \cite{zhou2019continuity} to represent rotations for effectively learning rotation quantities. Leveraging the scene geometry $\mathcal{S}$, sparse tracking signals $\mathbf{p}^{1:N}$ and the periodic alignment features $\mathbf{f}^{1:N}$, we obtain the final clean motion $\mathbf{x}^{1:N}_0$ by a conditional diffusion model, given initial noisy sample $\tilde{\mathbf{x}}^{1:N}$. We depict our full-body motion generation pipeline in Figure \ref{fig:pipeline}. 

\subsection{Motion Prior for $\text{S}^2$Fusion Initialization}
To circumvent the limited volume of scene-motion datasets and generate more diverse and realistic motions, we propose to draw initial motion distribution from a pre-trained motion prior for the reverse diffusion process \cite{lyu2022accelerating}. We build a VAE-based generative model that resembles an encoder-decoder architecture and trained on large-scale motion dataset AMASS \cite{mahmood2019amass}.

Once the VAE-based generative model is trained, we can sample initial motion unaware of the scene by conditioning on the tracking signals, 
\begin{equation}
    \tilde{\mathbf{x}}^{1:N} = f_{\phi}(z, \mathbf{p}^{1:N}),\quad z \sim \mathcal{N}(\mathbf{0}, \mathbf{I}),
\end{equation}
the sampled motion $\tilde{\mathbf{x}}^{1:N}$ is then refined by our conditional diffusion process Sec. \ref{subsec:diffusion}.

Compared to Gaussian noise, the pre-trained generative prior captures the complex structure of the motion manifold, therefore improving the sample quality. Meanwhile, by jump-starting to a noisy motion (as compared to standard DDPM settings which require $T=1000$ denoising steps \cite{lyu2022accelerating}), we can also achieve accelerated inference speed.

\subsection{Condition Acquisition for $\text{S}^2$Fusion}
\label{subsec:condition}

The overall conditioning input to our diffusion model is 
\begin{equation}
    \mathbf{c} = (\mathbf{p}^{1:N}, \mathbf{f}^{1:N}, \mathbf{E}_{\mathcal{S}}),
\end{equation}
consists of raw tracking signals, periodic alignment features, and encoded scene features. We introduce the acquisition of the conditioning inputs as follows.

\noindent\textbf{Scene conditioning.} During human-scene interaction, only a small region around the human provides meaningful contextual information. Hence we crop a $2m \times 2m \times 2m$ bounding box $\mathcal{B}_{\mathcal{S}}$ around the human, given the global translation measured by HMD. The cropped point cloud is then fed into a scene encoder \cite{qi2017pointnet++} $h_{\theta}$, obtaining the scene feature vector $E_{\mathcal{S}}$.

\noindent \textbf{Periodic motion feature.} Though the observed tracking signals consist of varying sources with different units and scales, i.e., rotation from the gyroscope, and position from the infrared optical sensor, the changes in these signals reflect the temporal movement pattern of the underlying motions. As suggested by \cite{starke2022deepphase}, full-body movements can happen as a composition of multiple local periodic movements, and these movements can be decomposed into multiple latent channels that capture the non-linear periodicity of different body segments, which are aligned in time. To effectively extract the temporal alignment feature of these signals, we resort to the frequency domain and use a periodic autoencoder (PAE) \cite{starke2022deepphase} $g_\tau$ to compute the frequency domain parameters \text{frequency} $\mathbf{F}$, \text{amplitude} $\mathbf{A}$, \text{offset} $\mathbf{B}$ and \text{phase shift} $\mathbf{S}$. Intuitively, the phase shift indicates the time-alignment of the motions, while the amplitude resembles the momentum.

\begin{figure}
    \centering
    \resizebox{\columnwidth}{!}{\includegraphics{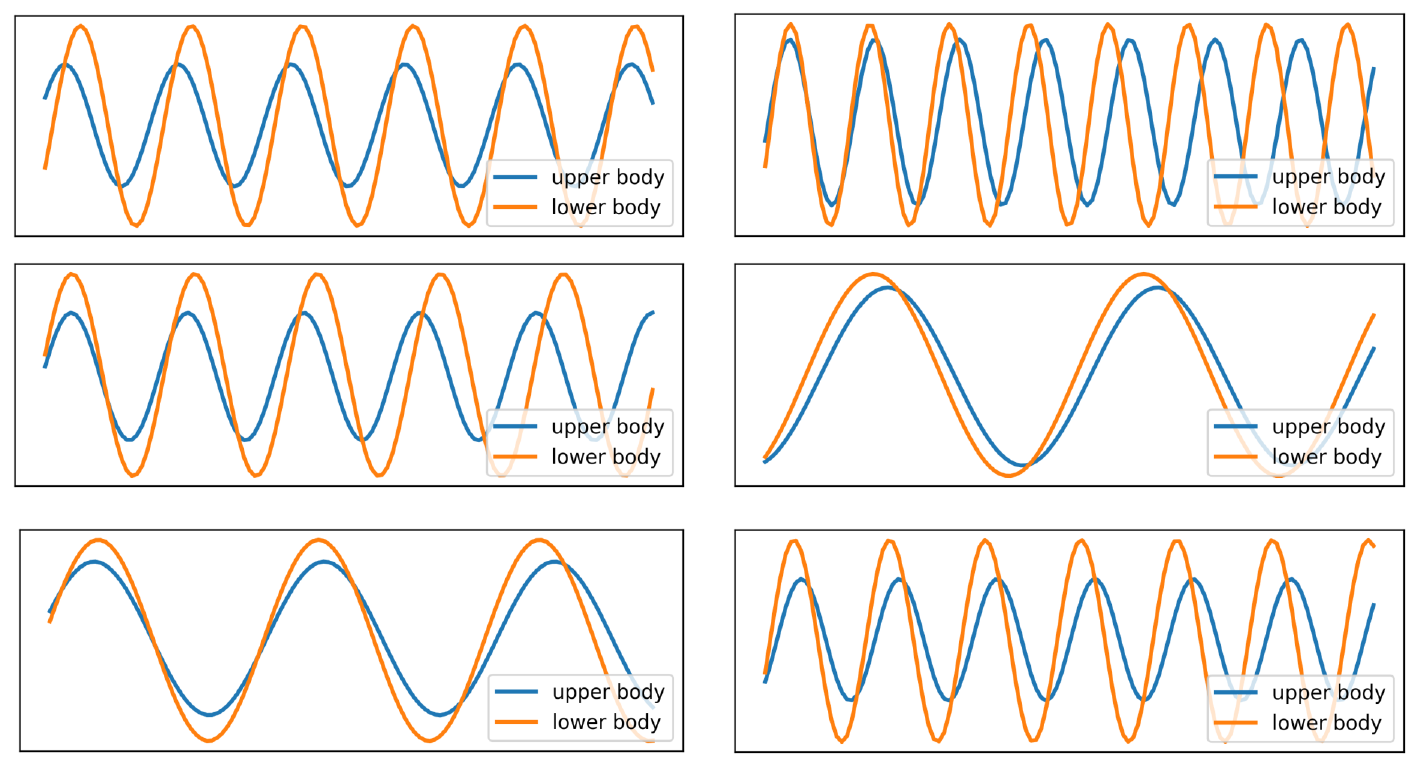}}
    \caption{A visualization of the periodic motion features of the upper and lower body extracted from randomly selected motion sequences in AMASS\cite{mahmood2019amass}. The \text{phase shift} of the sinusoidal functions indicates the time-alignment of the upper and lower body motions, while the \text{amplitude} resembles the momentum. It can be shown that the periodic motion features of the upper body are correlated with that of the lower body.}
    \label{fig:phase}
\end{figure}

The PAE $g_\tau$ consists of a combination of 1D convolution and a fast Fourier transformation (FFT) layer to compute the amplitudes $\mathbf{A}$, offsets $\mathbf{B}$, and frequencies $\mathbf{F}$ of a temporal signal $\mathbf{p}^{1:N}$,
\begin{equation}
    [\mathbf{A}, \mathbf{B}, \mathbf{F}]  = \text{FFT}(\text{Conv}(\mathbf{p}^{1:N})).
\end{equation}
\label{formula:fft}
The phase-shifts $\mathbf{S}$ are obtained by a separate fully-connected network,
\begin{equation}
    (s_x, s_y) = \text{FC}(\text{Conv}(\mathbf{p}^{1:N})), \mathbf{S} = \arctan(s_y, s_x).
\end{equation}
\label{formula:phase}
After extracting parameters in the frequency domain, we can reconstruct a smoothed periodic alignment feature in the temporal domain,
\begin{equation}
    \mathbf{f}_t = \mathbf{A} \cdot \sin(2\pi \cdot (\mathbf{F} \cdot t - \mathbf{S})) + \mathbf{B},
\end{equation}
where $t\in \{1,\dots, N\}$. The reconstructed feature in essence is a multi-resolution sinusoidal function, which summarizes the alignment of full-body motion in both time and space. \\

\noindent \textbf{Sparse tracking signals.} We take the position and rotation of the head and left/right hands as input signals, and compute angular and linear velocities as extra input following previous work \cite{jiang2022avatarposer}.  

\subsection{Conditional Denoiser of $\text{S}^2$Fusion}
\label{subsec:diffusion}
Diffusion models consist of a forward diffusion process and a reverse diffusion process. To model a distribution $\mathbf{x}_0 \sim q(\mathbf{x}_0)$, the forward process follows a Markov chain of $T$ steps which progressively adds Gaussian noise and produces a series of time-dependent distributions $q(\mathbf{x}_t | \mathbf{x}_{t-1})$. Formally,

\begin{equation}
    q(\mathbf{x}_{t}|\mathbf{x}_{t-1}) = \mathcal{N}(\mathbf{x}_t; \sqrt{1-\beta_t} \mathbf{x}_{t-1}, \beta_t \mathbf{I}),
\end{equation}

\begin{equation}
     q(\mathbf{x}_{1:T} | \mathbf{x}_{0}) = \prod_{t=1}^{T} q(\mathbf{x}_{t} | \mathbf{x}_{t-1}),
\end{equation}
where $\beta_t \in (0,1)$ is the variance schedule at timestep $t$.

To generate full-body motion, we model the conditioned motion generation problem by the reverse diffusion process, which gradually cleans corrupted signal $\mathbf{x}_T$. Instead of predicting residual Gaussian noise $\epsilon_t$ added through each step $t$, we predict the signal $\hat{\mathbf{x}}$ itself following \cite{tevet2022human} with the \textit{simple objective},

\begin{equation}
    \mathcal{L}_{\text{simple}} = \mathbb{E}_{t \sim [1,T]} \big\| G(\mathbf{x}_{t}^{1:N}, t, \mathbf{c}) - \mathbf{x}_0 \big\|
\end{equation}
\label{formula:simple}
This iterative process at timestep $t$ can be formulated as 
\begin{equation}
        \mathbf{x}_{t-1}^{1:N} = \sqrt{\bar{\alpha}_{t-1}} G(\mathbf{x}_{t}^{1:N}, t, \mathbf{c}) + \sqrt{1 - \bar{\alpha}_{t-1}}\epsilon,
\end{equation}
where $G$ is a network that learns to generate clean motion $\hat{\mathbf{x}}^{1:N}$ at timestep $t$, $\bar{\alpha}_t = \prod_{i=1}^{t} (1 - \beta_{i})$ and $\epsilon \sim \mathcal{N}(\mathbf{0}, \mathbf{I})$ is the injected Gaussian noise. The denoising network $G$ is trained with $\mathcal{L}_{\text{simple}}$ and geometric loss \cite{tevet2022human} that regulates the generated motion to lie on motion manifold, 
\begin{equation}
    \mathcal{L}_{\text{train}} = \mathcal{L}_{\text{simple}} + \mathcal{L}_{\text{geometric}}, 
\end{equation}

\begin{equation}
    \mathcal{L}_{\text{geometric}} = \| \text{FK}(\hat{\mathbf{x}}) - \text{FK}(\mathbf{x}_0) \|,
\end{equation}
where the $\text{FK}(\cdot)$ denotes the forward kinematic process that converts joint rotations into positions.

More details about the training details of the VAE-based motion prior and periodic autoencoder are provided in the supplementary.

\subsection{$\text{S}^2$Fusion Sampling with Lower Body Motion Regularization}
\label{subsec:guiding} 
Given the sparse tracking signals of the \textit{upper body}, it is already possible to reconstruct the upper body motion faithfully \cite{jiang2022avatarposer, du2023agrol, zheng2023realistic}. However, the lack of lower body tracking signals makes the generated leg motions often possess unrealistic behavior and are inconsistent with the upper body. Special treatment to the lower body is required to generate more realistic motions. To tackle the challenge presented by the absence of observations, we designed two loss functions that regularize the leg motions which \textit{do not require any tracking signals from the lower body}. We incorporate the designed loss functions in the sampling procedure of our $\text{S}^2$Fusion, enabling flexible control of the lower body motions. 

\input{algorithm}

\noindent\textbf{Scene-penetration loss} resolves the incurring of implausible human-scene penetration,


\begin{equation}
    \ell_{\text{penetration}}(\mathbf{x}_0) = \sum_{i \in \mathcal{C}} \sum_{b \in \text{KNN}(\mathbf{x}_{0, i}, k)} \max(r - \|\mathbf{x}_{0, i} - b\|, 0), 
\end{equation}
where $\mathcal{C}$ is the set of joints that we want to avoid contact with the scene, $k$ is the number of neighbors to the KNN query, and points in the scene within radius $r$ are considered to be in contact. We empirically set $\mathcal{C}$ as the left/right ankles and the left/right knees; we set $r=0.02$m and $k=4$ in our evaluation.

\noindent\textbf{Phase-matching loss} forces the upper body and the lower body to move in a coherent manner,

\begin{equation}
    \ell_{\text{phase}}(\mathbf{x}_0) = \| P_{\text{upper}} - P_{\text{lower}} \|,
\end{equation}
where $P_{\text{upper}}$ is the phase feature computed using the amplitude and phase shift as defined in Sec. \ref{formula:phase},

\begin{equation}
    P_{\text{upper}} = [\sin (2\pi\cdot\mathbf{S}), \cos (2 \pi \cdot \mathbf{S}), \mathbf{A}].
\end{equation}
The effectiveness of the phase-matching loss can be justified by the observation that the upper and lower bodies always move in a coordinated way: humans tend to synchronize the upper and lower body movements to achieve balance during everyday activities such as walking, running, dancing, and so on. The movement of the \textit{upper body} provides a valuable cue to the movement of the \textit{lower body}. We plot the periodic motion features of the upper and lower body in Figure \ref{fig:phase}. We observe there is a clear correlation between the upper and lower body movements.

To compute the phase feature $P_{\text{lower}}$ of the lower body, we select the anchor joints that can represent the motion of the lower body: the pelvis and left/right ankles, resembling the tracking signals sent from the head and left/right hands. Then we collect the position, rotation, linear velocity, and angular velocity of the generated lower body motions, following the same procedure in Sec. \ref{subsec:condition}. We then compute the phase shift and amplitude parameter and obtain $P_{\text{lower}}$.

\noindent \textbf{Sampling with guidance from loss functions.} The overall loss guidance in the sampling stage of $\text{S}^2$Fusion is as follows,
\begin{equation}
    \ell_{\text{sample}} = 
    \alpha \cdot \ell_{\text{penetration}} + \beta \cdot \ell_{\text{phase}},
\end{equation}
where $\alpha$ and $\beta$ are scaling factors controlling the strength of applying guidance.

After obtaining a clean motion sample $\hat{\mathbf{x}}^{1:N}_0$ as in Sec. \ref{subsec:diffusion}, we guide the sampling process by injecting the gradient of $\ell_{\text{sample}}$ to regularize the motion of lower body,
\begin{equation}
    \bar{\mathbf{x}}_{0}^{1:N} \gets \hat{\mathbf{x}}^{1:N}_0 - \eta \nabla_{\hat{\mathbf{x}}^{1:N}_0} \ell_{\text{sample}}(\hat{\mathbf{x}}^{1:N}_0).
\end{equation}
The full pipeline of the sampling procedure is summarized in Algorithm 1. We chose $\alpha=0.1, \beta=0.01, \eta=1$ empirically.

\section{Experiments}
\label{sec:experiments}
We first evaluate and compare our method to others \cite{jiang2022avatarposer, du2023agrol, zheng2023realistic} in commonly used metrics for motion reconstruction. Then we conduct ablation studies to showcase the effectiveness of the design choices of our model. 

\input{tables/main}

\begin{figure*}
    \centering
    \resizebox{2\columnwidth}{!}{\includegraphics{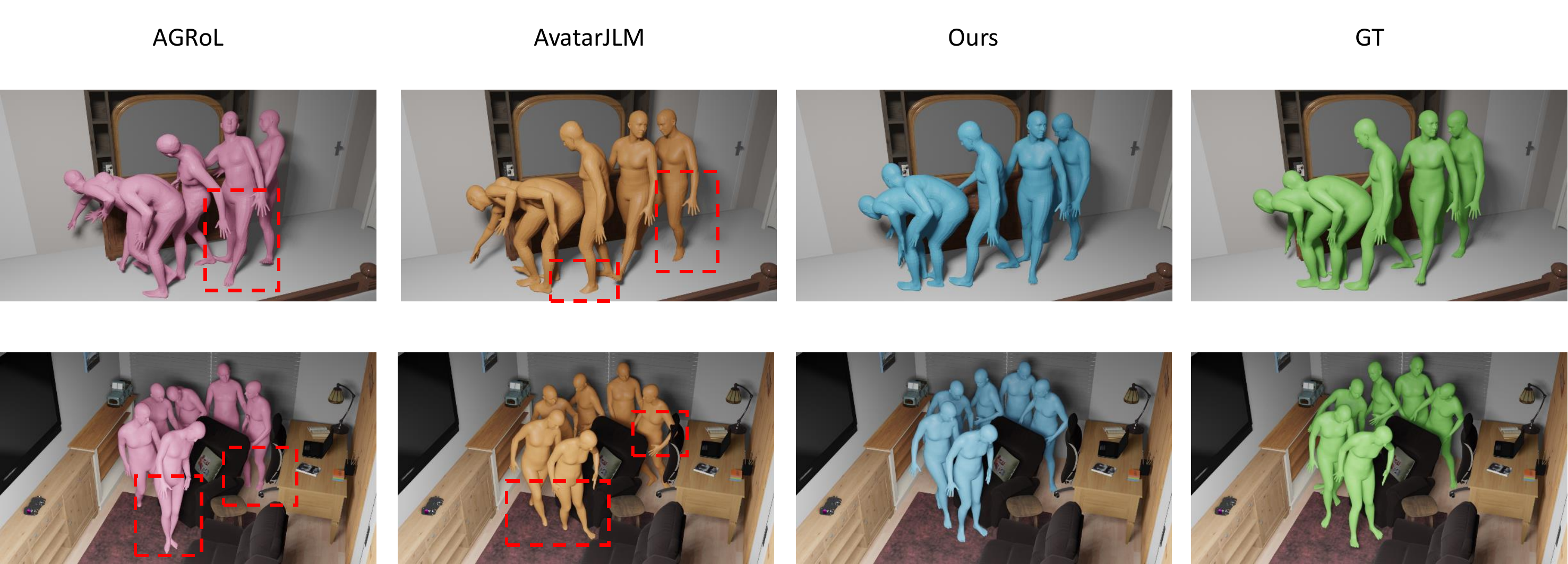}}
    \caption{Qualitative results on the CIRCLE \cite{Araujo_2023_CVPR} dataset. We show the results of two motion sequences in different scenes and highlight the implausible motions in the red box. It can be shown that our method generates more correlated leg motions and avoids scene penetration as much as possible.}
    \label{fig:exp}
\end{figure*}

\subsection{Experiment Setup}
\subsubsection{Datasets}
We consider two motion-scene datasets to train and evaluate our method. Both datasets consist of rich human-scene interactions and provide fine-grained scene meshes.

\noindent\textbf{CIRCLE} \cite{Araujo_2023_CVPR} contains 10 hours of full-body motion in 9 diverse scenes paired with egocentric information of the environment The CIRCLE dataset is created by integrating real-world motion capture sequence into a high-quality artist-created virtual environment. We randomly split the training and testing data by a 70:30 ratio, and discarded sequences shorter than 1.0s.

\noindent\textbf{GIMO} \cite{zheng2022gimo} consists of body pose sequences, scene scans, and eye gaze information. The dataset is collected by using Hololens and IMU-based motion capture suits for motion acquisition; and iPhone 12 for scene scanning. 

\subsubsection{Evaluation Metrics}
We evaluate the performance of our method against others using the commonly used evaluation metrics, which measure motion estimation accuracy, motion smoothness, and scene awareness, respectively.

\noindent \textbf{MPJRE} (mean per joint rotation error) measures the average relative rotation error of all body joints in $\deg$. \textbf{MPJPE} (mean per joint position error) measures the pose accuracy of each frame in $mm$. \textbf{MPJVE} (mean per joint velocity error) measures the average velocity error of all body joints in $mm/s$. \textbf{Jitter} measures the average jerk, which is the time derivative of acceleration, of all body joints and reflects the smoothness of motion. \textbf{FS} represents the accumulated drift of foot joints during contact, it is computed by the average horizontal displacement between the grounding feet in adjacent frames.

We also conduct per-body part evaluations, as \textbf{Hand PE}, \textbf{Upper PE} and \textbf{Lower PE} denote the hand position error, upper body position error and lower body position error, respectively.

\subsection{Comparison}
We compare our method with the state-of-the-art methods, including AvatarPoser \cite{jiang2022avatarposer}, AGRoL \cite{du2023agrol}, and AvatarJLM \cite{zheng2023realistic} on CIRCLE \cite{Araujo_2023_CVPR} and GIMO \cite{zheng2022gimo} datasets. For a fair comparison, we re-train these methods on CIRCLE and GIMO until convergence.

We show our quantitative results in Table \ref{table:main} and Table \ref{table:upper-lower}. Compared to other methods, our method achieves the best MPJPE and FS among others in both CIRCLE and GIMO datasets, demonstrating the effectiveness of incorporating scene information to resolve the sparse-to-dense ambiguities and to generate much more accurate motions. Moreover, our $\text{S}^2$Fusion model improves the smoothness in motion generation, as showcased by the \textit{MPJVE}, \textit{Jitter} and \textit{FS} metrics in comparison to other methods. It could be also observed that the estimation quality of the lower body motion is significantly better than other methods by a large margin. The superiority of our method is also evident in the qualitative results shown in Figure \ref{fig:exp}. By visualizing the pose trajectory in different scenes, we demonstrate that our method generates accurate and smooth motions; being able to incorporate scene information as inputs, the motion generated by our method incurred fewer scene penetrations.

\input{tables/upperlower}

\subsection{Ablation Study}
\label{subsec: ablation}
To further investigate the effectiveness of design choices in $\text{S}^2$Fusion, we perform an ablation study to study the effect of different components. We first ablate the effectiveness of our model components, including the benefit of introducing the scene as an extra modality, and the use of pre-trained motion prior and periodic autoencoder. Then we study the effectiveness of our specially designed loss function during the loss-guided sampling process.

\input{tables/ablation-model}

\subsubsection{Effectiveness of $\text{S}^2$Fusion Modules Design}
To validate the effectiveness of incorporating a pre-trained motion prior and periodic autoencoder, as well as introducing the scene as an extra modality to the input of our model, we compare our method against four alternatives, as shown in Table \ref{table:ablation-circle} and Table \ref{table:ablation-gimo}. The result tables clearly demonstrate introducing the extra scene modality can significantly improve the motion estimation quality when only sparse tracking signals from the upper body are available. It verified our claim that scene information can greatly reduce the one-to-many ambiguities presented in this sparse-to-dense problem. We also observed that drawing the initial sample from the pre-trained motion prior can circumvent the limited data volume issue in common motion-scene datasets, and solely leveraging motion prior can achieve a big improvement in motion smoothness and estimation accuracy. The periodic autoencoder can also improve the motion estimation quality as it can effectively extract the spatial-temporal alignment of the motions from the sparse tracking signal inputs.

\subsubsection{Effectiveness of Loss-guided Sampling} 
We ablate the importance of incorporating our designed loss function in the loss-guided sampling process. We show the results compared against three alternatives in Table \ref{table:ablation-circle-loss} and Table \ref{table:ablation-gimo-loss}. We notice that although the scene-penetration loss and phase-matching loss can improve the motion estimation quality, they may introduce jittering motions into the generated motions. Compared to scene-penetration loss, incorporating phase-matching loss is more effective in producing accurate motions, coinciding with our observation that the phase features are composed of time-alignment features and feature vector resembles momentum. The scene-penetration loss has merit on its own as it can avoid falsely contacting the ground, especially in the CIRCLE dataset.

\input{tables/ablation-score}

\section{Conclusion}
Our work addresses the pivotal challenge of estimating full-body human motion in 3D scenes from sparse tracking signals, a critical aspect for advancing AR/VR applications. To tackle this problem, we propose $\text{S}^2$Fusion that integrates \underline{S}cene and sparse \underline{S}ignals through a conditional Dif\underline{Fusion} model. $\text{S}^2$Fusion first captures spatial-temporal relations in sparse signals using a periodic autoencoder, generating time-alignment feature embeddings as additional inputs. Employing conditional diffusion and drawing initial motion from a pre-trained prior, $\text{S}^2$Fusion effectively fuses scene geometry and sparse tracking signals to generate full-body scene-aware motions. 
To improve plausibility and coherence, $\text{S}^2$Fusion incorporates a specially designed scene-penetration loss and phase-matching loss, providing guidance for the sampling procedure. These losses effectively regularize lower-body motion, even in the absence of tracking signals. Extensive experiments demonstrate that $\text{S}^2$Fusion outperforms state-of-the-art methods by a large margin. 
Extending $\text{S}^2$Fusion for human motion estimation in more complex scenarios by systematically integrating comprehensive physically plausible constraints is under consideration for our future work. 
\label{sec:conclusion}

\section*{Acknowledgement}
This work was supported by Shanghai Local College Capacity Building Program (23010503100), NSFC (No.62303319), Shanghai Sailing Program (22YF1428800, 21YF1429400), Shanghai Frontiers Science Center of Human-centered Artificial Intelligence (ShangHAI), MoE Key Laboratory of Intelligent Perception and Human-Machine Collaboration (ShanghaiTech University), and Shanghai Clinical Research and Trial Center.

%% file: algorithm.tex
\begin{algorithm}[!ht]
\DontPrintSemicolon
  
  \KwInput{sparse tracking signals $\mathbf{p}^{1:N}$, scene point cloud $\mathcal{S}$}
  \KwOutput{full-body motion $\mathbf{x}^{1:N}_{0}$}
  \tcc{preprocessing}
  
  $z \sim \mathcal{N}(\mathbf{0}, \mathbf{I})$
  
  Draw initial noisy motion from pre-trained prior $\tilde{\mathbf{x}}^{1:N} \sim f_\phi(z, \mathbf{p}^{1:N})$

  Extract periodic motion feature $\mathbf{f}^{1:N} := g_{\tau}(\mathbf{p}^{1:N})$

  Extract scene feature $\mathbf{E}_{\mathcal{S}} := h_\theta (\mathcal{B}_{\mathbf{S}})$

  Form conditioning input $\mathbf{c} := (\mathbf{p}^{1:N}, \mathbf{f}^{1:N}, \mathbf{E}_{\mathcal{S}})$

  \tcc{reverse diffusion process}
  $\mathbf{x}_{T}^{1:N} := \tilde{\mathbf{x}}^{1:N}$
  
  \For{$t=T, \dots, 1$}
  {
    $\hat{\mathbf{x}}^{1:N}_0 := G(\mathbf{x}_{t}^{1:N}, t, \mathbf{c})$ \tcp*{predict clean signal}

    $\bar{\mathbf{x}}^{1:N}_0 := \hat{\mathbf{x}}^{1:N}_0 - \eta \nabla_{\hat{\mathbf{x}}^{1:N}_0} \ell (\hat{\mathbf{x}}^{1:N}_0)$ \tcp*{loss guidance}

    $\mathbf{x}_{t-1}^{1:N} := \sqrt{\bar{\alpha}_{t-1}} \bar{\mathbf{x}}_{0}^{1:N} + \sqrt{1 - \bar{\alpha}_{t-1}} \mathbf{\epsilon}$
  }

  \textbf{return} $\mathbf{x}^{1:N}_0$

\caption{Sampling procedure of $\text{S}^2$Fusion for full-body human motion estimation.} 
\label{alg:cap}
\end{algorithm}
\label{algo}

%% file: tables/main.tex
\begin{table*}
\centering
\resizebox{2\columnwidth}{!}{
\begin{tabular}{ccccccccccc}
\hline
& \multicolumn{4}{c}{GIMO \cite{zheng2022gimo}} & \multicolumn{4}{c}{CIRCLE \cite{Araujo_2023_CVPR}} \\ 
\hline
\bfseries Method & \textbf{MPJRE} $\downarrow$ & \textbf{MPJPE} $\downarrow$ & \textbf{MPJVE} $\downarrow$ & \textbf{Jitter} $\downarrow$ & \textbf{FS} $\downarrow$ & \textbf{MPJRE} $\downarrow$ & \textbf{MPJPE} $\downarrow$ & \textbf{MPJVE} $\downarrow$ & \textbf{Jitter} $\downarrow$ & \textbf{FS} $\downarrow$ \\ 
\hline
AvatarPoser \cite{jiang2022avatarposer} & 7.02 & 91.3 & 324.0 & 16.4 & 2.04 & 2.68 & 30.5 & 184.6 & 11.5 & 2.11 \\
AGRoL \cite{du2023agrol} & 6.58 & 88.6 & 269.4 & 12.5 & 1.70 & 2.62 & 28.7 & 141.1 & 8.2 & 1.72 \\
AvatarJLM \cite{zheng2023realistic} & 4.95 & 70.7 & 258.1 & 10.7 & 1.41 & 2.49 & 24.6 & 128.5 & 7.0 & 1.53 \\
\hline
Ours & \textbf{4.65} & \textbf{57.8} & \textbf{235.7} & \textbf{10.1} & \textbf{1.39} & \textbf{2.32} & \textbf{19.2} & \textbf{117.6} & \textbf{5.8} & \textbf{1.48} \\
\hline
\end{tabular}
}   
\caption{Full-body motion estimation results evaluated on GIMO \cite{zheng2022gimo} and CIRCLE \cite{Araujo_2023_CVPR}}
\label{table:main}
\end{table*}

%% file: tables/upperlower.tex
\begin{table*}[t]
\centering

\begin{tabular}{cccccccccccc}
\hline
& \multicolumn{3}{c}{\textbf{GIMO} \cite{zheng2022gimo}} & \multicolumn{3}{c}{\textbf{CIRCLE} \cite{Araujo_2023_CVPR}} \\ 
\hline
\bfseries Method & \textbf{Hand PE} & \textbf{Upper PE} & \textbf{Lower PE} & \textbf{Hand PE} & \textbf{Upper PE} & \textbf{Lower PE} \\
\hline
AvatarPoser \cite{jiang2022avatarposer} & 36.8 & 40.1 & 198.6 & 11.4 & 15.3 & 83.3 \\
AGRoL \cite{du2023agrol} & 23.7 & 29.8 & 163.9 & 9.1 & 14.5 & 77.4 \\ 
AvatarJLM \cite{zheng2023realistic} & 25.0 & 33.4 & 132.6 & 9.3 & 15.8 & 69.1 \\ 
\hline
Ours & \textbf{23.1} & \textbf{28.7} & \textbf{107.9} & \textbf{8.8} & \textbf{12.2} & \textbf{57.3} \\
\hline
\end{tabular}
\caption{More metrics comparison with AvatarPoser \cite{jiang2022avatarposer}, AGRoL \cite{du2023agrol}, and AvatarJLM \cite{zheng2023realistic} on GIMO \cite{zheng2022gimo} and CIRCLE \cite{Araujo_2023_CVPR}. We show the results of comparing the hand, upper body, and lower body reconstruction quality.}
\label{table:upper-lower}
\end{table*}

%% file: tables/ablation-model.tex
\begin{table}[t]

\resizebox{\columnwidth}{!}{
\begin{tabular}{ cccccccc } 
\hline
\multicolumn{3}{c}{\textbf{Components}} &  &  &  &  & \\
\cline{1-3} 
\textbf{MP} & \textbf{Scene} & \textbf{PAE} & \textbf{MPJPE} & \textbf{MPJVE} & \textbf{Jit.} & \textbf{FS} \\
\hline
$\times$ & $\times$ & $\times$ & 26.2 & 135.9 & 7.9 & 1.65 \\
$\times$ & $\checkmark$ & $\times$ & 22.7 & 128.1 & 7.1 & 1.58 \\
$\checkmark$ & $\checkmark$ & $\times$ & 20.9 & 120.5 & 6.0 & 1.51 \\ 
$\times$ & $\checkmark$ & $\checkmark$ & 21.8 & 125.3 & 6.6 & 1.52  \\
\hline
$\checkmark$ & $\checkmark$ & $\checkmark$ & \textbf{19.2} & \textbf{117.6} & \textbf{5.8} & \text{1.48}  \\
\hline
\end{tabular}
}
\caption{Ablation on various components of our model on CIRCLE. \textbf{MP} denotes the pre-trained motion prior, \textbf{Scene} indicates whether the model receives the scene information as an extra input, and \textbf{PAE} denotes the periodic autoencoder.}
\label{table:ablation-circle}
\end{table}

\begin{table}[t]

\resizebox{\columnwidth}{!}{
\begin{tabular}{ cccccccc } 
\hline
\multicolumn{3}{c}{\textbf{Components}} &  &  &  &  & \\
\cline{1-3} 
\textbf{MP} & \textbf{Scene} & \textbf{PAE} & \textbf{MPJPE} & \textbf{MPJVE} & \textbf{Jit.} & \textbf{FS} \\
\hline
$\times$ & $\times$ & $\times$ & 76.6 & 264.9 & 11.7 & 1.68  \\
$\times$ & $\checkmark$ & $\times$ & 68.1 & 257.4 & 11.2 & 1.67 \\
$\checkmark$ & $\checkmark$ & $\times$ & 60.3 & 243.6 & 10.3 & 1.46 \\ 
$\times$ & $\checkmark$ & $\checkmark$ & 65.9 & 249.3 & 10.4 & 1.52 \\
\hline
$\checkmark$ & $\checkmark$ & $\checkmark$ & \textbf{57.8} & \textbf{235.7} & \textbf{10.1} & \textbf{1.39} \\
\hline
\end{tabular}
}
\caption{Ablation on various components of our model on GIMO. \textbf{MP} denotes the pre-trained motion prior, \textbf{Scene} indicates whether the model receives the scene information as an extra input, and \textbf{PAE} denotes the periodic autoencoder.}
\label{table:ablation-gimo}
\end{table}

%% file: tables/ablation-score.tex
\begin{table}[t]
\resizebox{\columnwidth}{!}{
\begin{tabular}{cccccc}
\hline
\multicolumn{2}{c}{\textbf{Loss fn.}} &  &  &  & \\
\cline{1-2}
$\ell_{\text{penetration}}$ & $\ell_{\text{phase}}$ & \textbf{MPJPE} & \textbf{MPJVE} & \textbf{Jit.} & \textbf{FS} \\ 
\hline
$\times$ & $\times$ & 20.6 & 119.1 & \textbf{5.6} & 1.43 \\
$\checkmark$ & $\times$ & 20.1 & 117.9 & 5.8 & \textbf{1.40} \\ 
$\times$ & $\checkmark$ & 19.8 & 118.5 & 6.1 & 1.50 \\ 
\hline
$\checkmark$ & $\checkmark$ & \textbf{19.2} & \textbf{117.6} & 5.8 & 1.48 \\
\hline
\end{tabular}
}
\caption{Ablation on the effect of our designed loss function during loss-guided sampling on CIRCLE.}
\label{table:ablation-circle-loss}
\end{table}

\begin{table}[t]
\resizebox{\columnwidth}{!}{
\begin{tabular}{cccccc}
\hline
\multicolumn{2}{c}{\textbf{Loss fn.}} &  &  &  & \\
\cline{1-2}
$\ell_{\text{penetration}}$ & $\ell_{\text{phase}}$ & \textbf{MPJPE} & \textbf{MPJVE} & \textbf{Jit.} & \textbf{FS} \\ 
\hline
$\times$ & $\times$ & 59.9 & 240.5 & 10.5 & 1.52 \\
$\checkmark$ & $\times$ & 58.3 & 239.1 & 10.4 & 1.58 \\ 
$\times$ & $\checkmark$ & \textbf{57.6} & 236.8 & 10.1 & 1.42 \\ 
\hline
$\checkmark$ & $\checkmark$ & 57.8 & \textbf{235.7} & \textbf{10.1} & \textbf{1.39} \\
\hline
\end{tabular}
}
\caption{Ablation on the effect of our designed loss function during loss-guided sampling on GIMO.}
\label{table:ablation-gimo-loss}
\end{table}

%% file: 12_appendix.tex
\section{Model details}

\begin{figure*}
    \centering
    \resizebox{2\columnwidth}{!}{\includegraphics{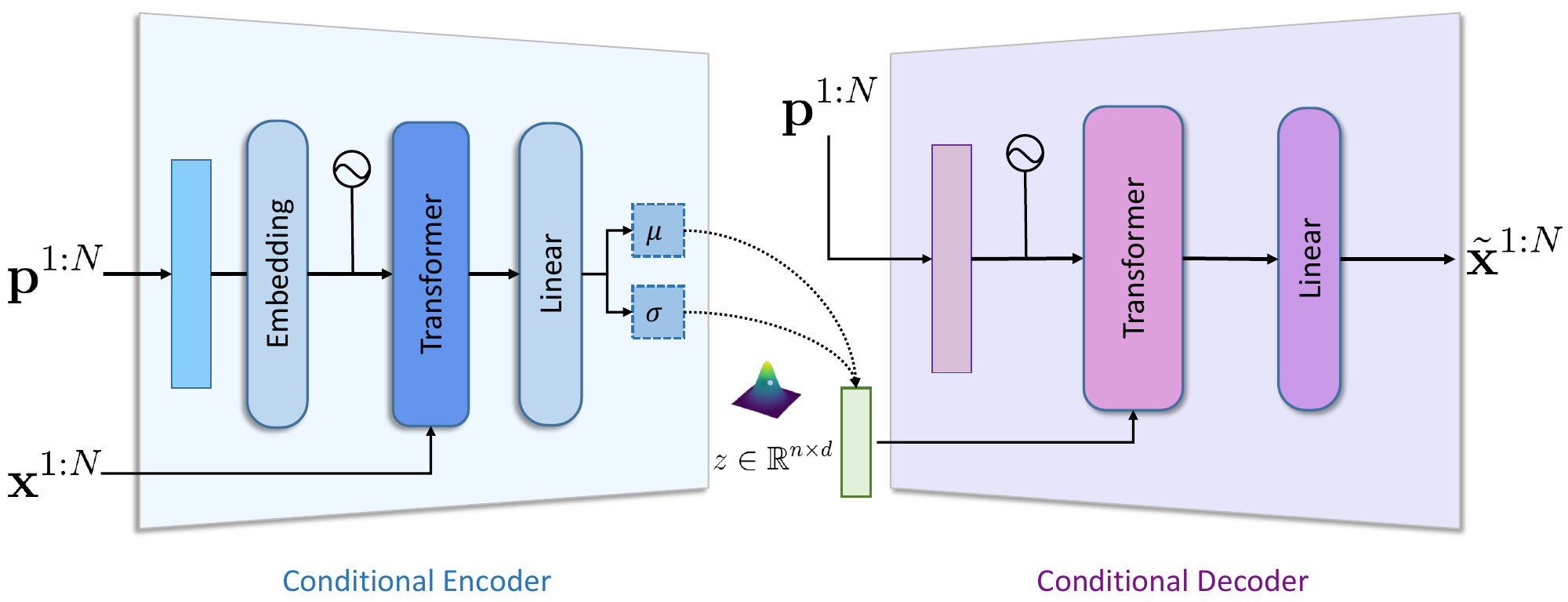}}
    \caption{The structure of our VAE-based motion prior, consists of encoder $\mathcal{E}$ and decoder $\mathcal{D}$.}
    \label{fig:supp-vae}
\end{figure*}

\begin{figure}
    \centering
    \resizebox{\columnwidth}{!}{\includegraphics{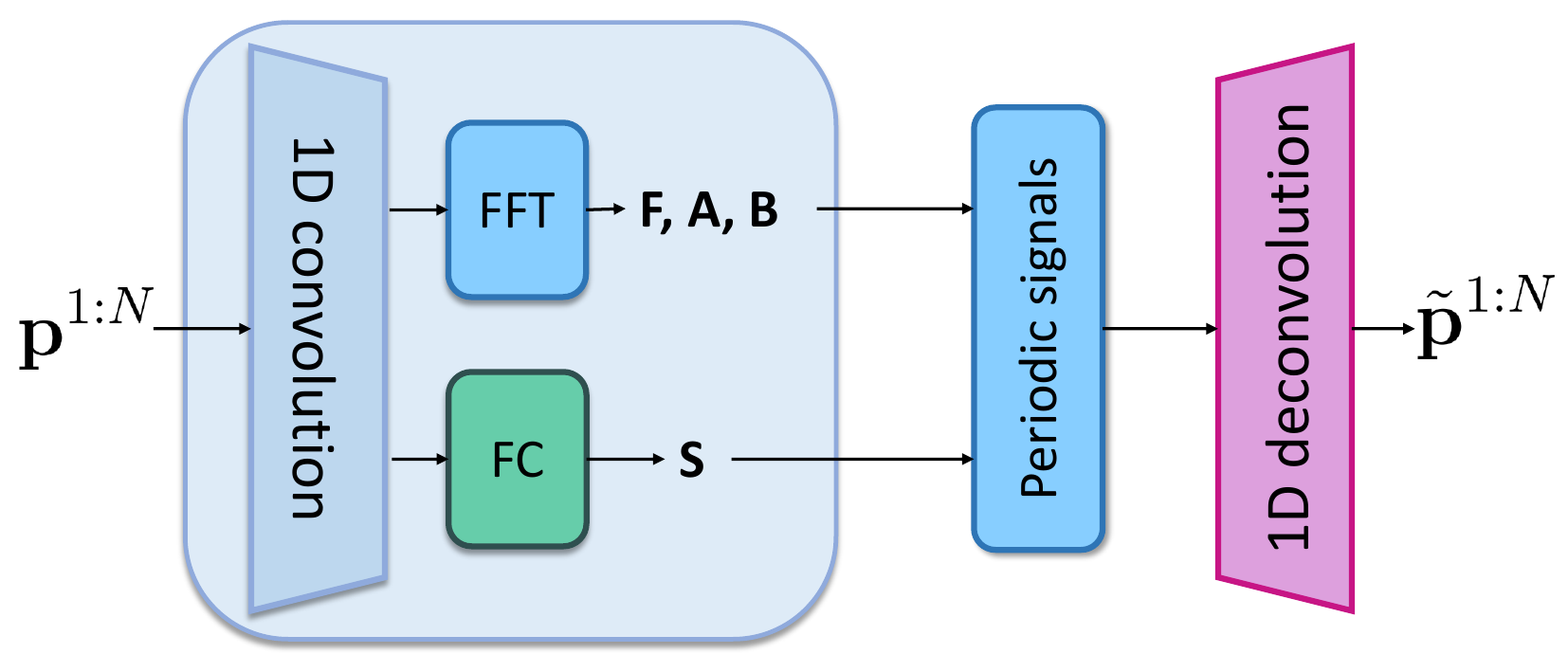}}
    \caption{The structure of the periodic autoencoder.}
    \label{fig:supp-phase}
\end{figure}

We first supplement our pipeline's omitted implementation details, including the architecture of VAE-based motion prior and periodic autoencoder, the pre-training of our motion prior and periodic autoencoder, and the hyperparameter settings.

\subsection{VAE-based motion prior}

Figure \ref{fig:supp-vae} shows the detailed structure of the VAE-based motion prior. Our VAE-based motion prior consists of a conditional encoder $\mathcal{E}$ and decoder $\mathcal{D}$.

Given a sequence of sparse tracking signals $\mathbf{p}^{1:N}$ and paired full-body motions $\mathbf{x}^{1:N}$, the conditional encoder outputs a latent code,
\begin{equation}
    z = \mathcal{E}(\mathbf{x}^{1:N} | \mathbf{p}^{1:N}).
\end{equation}

The conditional decoder $\mathcal{D}$ learns to reconstruct the full-body motions by given latent code $z$ and sparse tracking signals $\mathbf{p}^{1:N}$,
\begin{equation}
    \hat{\mathbf{x}}^{1:N} = \mathcal{D}(z | \mathbf{p}^{1:N}).
\end{equation}

The conditional encoder $\mathcal{E}$ is discarded during inference, and we only use the conditional decoder $\mathcal{D}$. \\

\noindent \textbf{Training objectives.}
The training objectives of the VAE-based motion prior is
\begin{equation}
    \mathcal{L}_{\text{VAE}} = \lambda_{\text{KL}} \cdot \mathcal{L}_{\text{KL}} + \lambda_{\text{recon}} \cdot \mathcal{L}_{\text{recon}} + \lambda_{\text{geometric}} \cdot \mathcal{L}_{\text{geometric}}.
\end{equation}

The KL divergence $\mathcal{L}_{\text{KL}}$ minimizes the distribution distance between the learned conditional distribution $p_{\mathcal{E}}(z|\mathbf{x}^{1:N}, \mathbf{p}^{1:N})$ and the standard Gaussian distribution $q(z) \sim \mathcal{N}(\mathbf{0}, \mathbf{I})$.

The reconstruction loss forces the model to learn an informative latent $z$ and be able to recover full-body motions from such latents,

\begin{equation}
    \mathcal{L}_{\text{recon}}(\hat{\mathbf{x}}^{1:N}, \mathbf{x}^{1:N}) = \left\|\hat{\mathbf{x}}^{1:N} - \mathbf{x}^{1:N}\right\|^2.
\end{equation}

The geometric loss $\mathcal{L}_{\text{geometric}}$ is the same as we mentioned in the text; it regularizes the generated motion to lie on the motion manifold.

We empirically set $\lambda_\text{KL} = 0.002$, $\lambda_{\text{recon}} = 1.0$, $\lambda_{\text{geometric}} = 0.5$ during training.

\begin{figure}
    \centering
    \resizebox{\columnwidth}{!}{\includegraphics{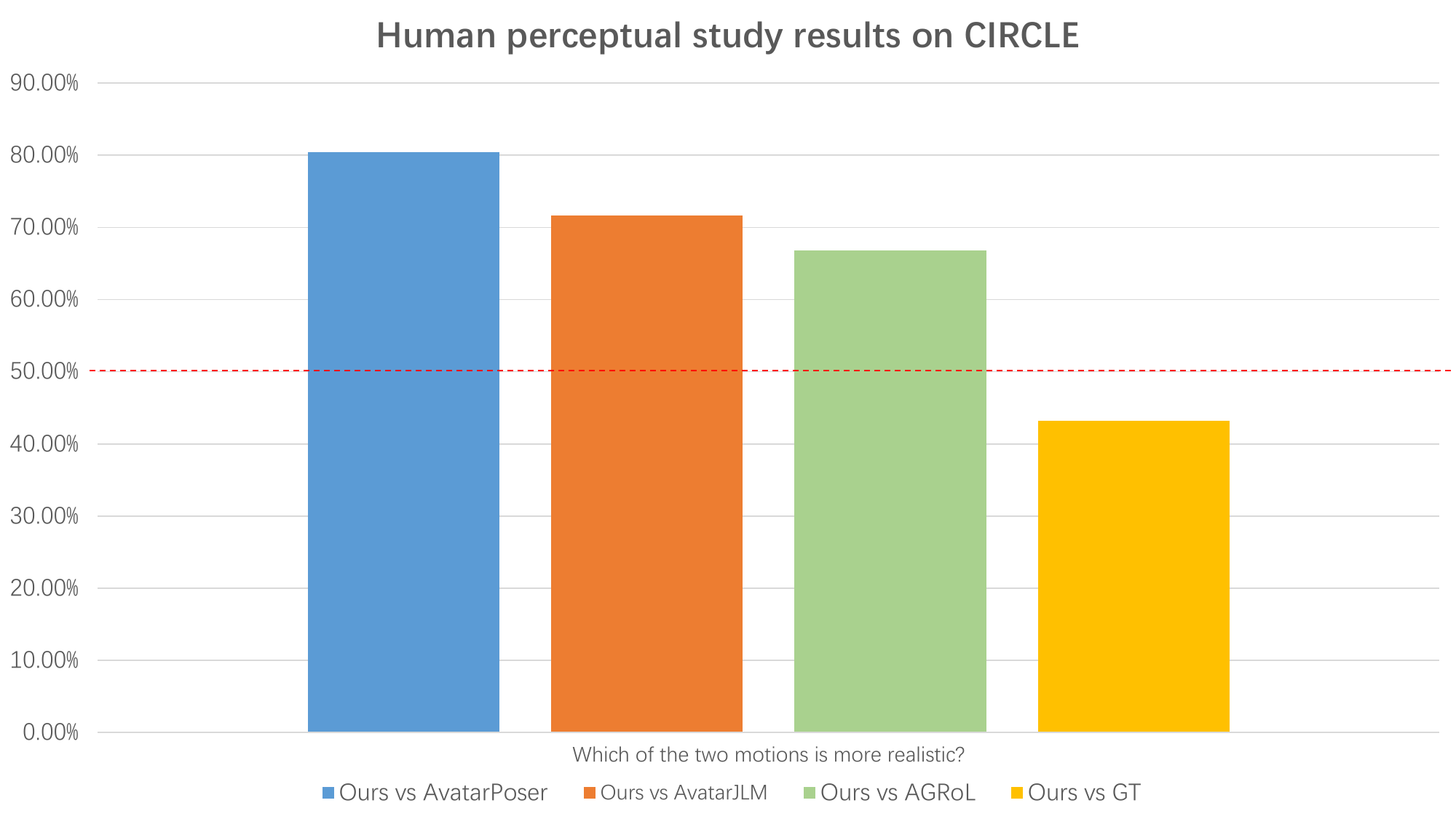}}
    \caption{Human perceptual study results on the CIRCLE dataset.}
    \label{fig:supp-perceptual-circle}
\end{figure}

\begin{figure}
    \centering
    \resizebox{\columnwidth}{!}{\includegraphics{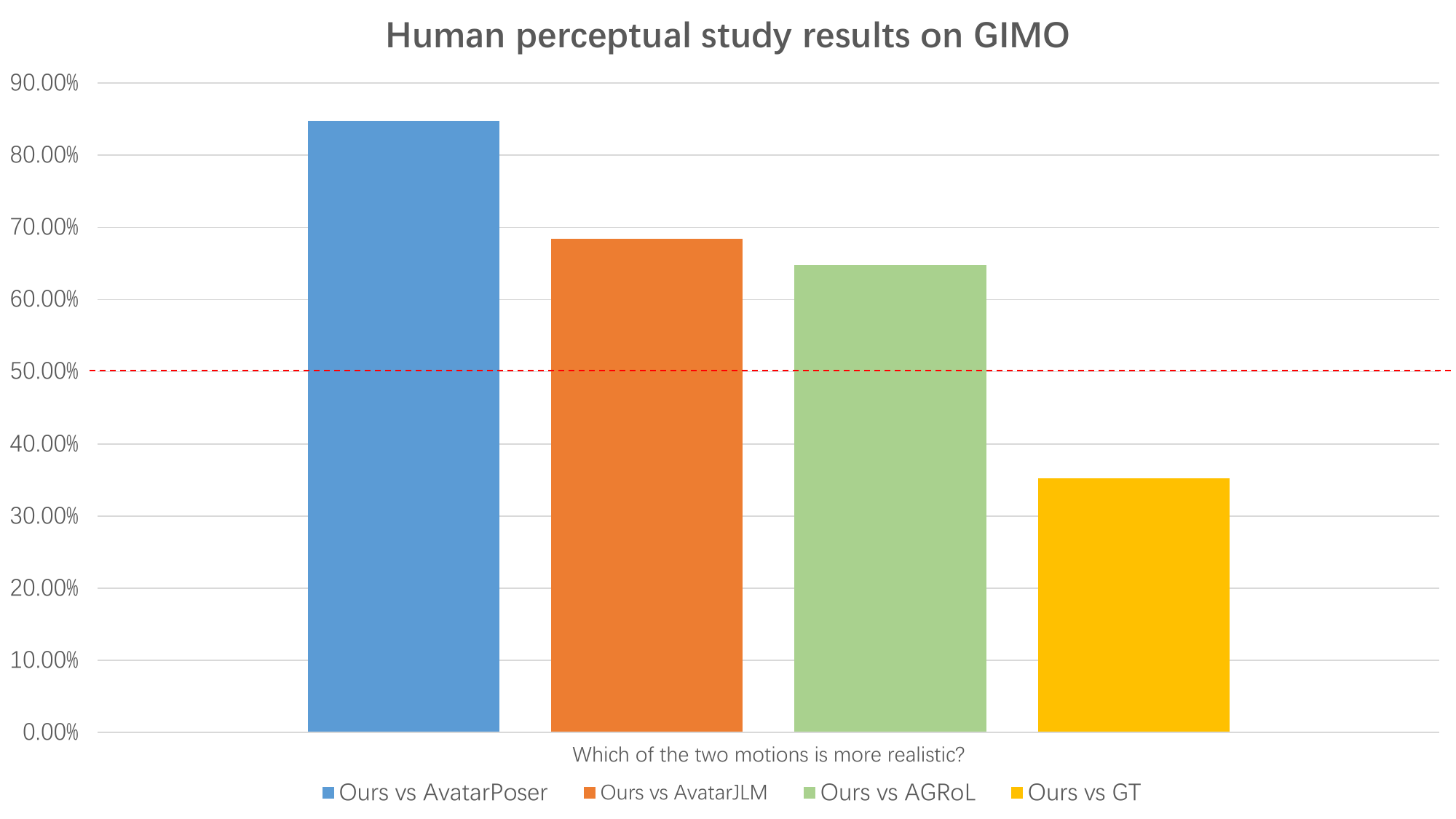}}
    \caption{Human perceptual study results on the GIMO dataset.}
    \label{fig:supp-perceptual-gimo}
\end{figure}

\subsection{Periodic autoencoder}
The full pipeline of the periodic autoencoder, shown in Figure \ref{fig:supp-phase}, consists of an encoder and decoder, whereas in the text we only discuss the encoder part.

Starting with encoded feature maps $\mathbf{f}^{1:N}$ in the temporal domain, we reconstruct the original tracking signals $\tilde{\mathbf{p}}^{1:N}$ by a 1D deconvolution,

\begin{equation}
    \tilde{\mathbf{p}}^{1:N} = \text{DeConv}(\mathbf{f}^{1:N}).
\end{equation}

The entire PAE is pre-trained using reconstruction loss,
\begin{equation}
    \mathcal{L}_{\text{PAE}} = \left\| \mathbf{p}^{1:N} - \tilde{\mathbf{p}}^{1:N} \right\|.
\end{equation}

During inference, we kept only the \text{encoder part} to extract the temporal periodic feature maps and the related phase features. 

\section{Implementation details}
The overall inputs to our model consist of encoded scene feature $\mathbf{E}_{\mathcal{S}} \in \mathbb{R}^{n}$, sparse tracking signals $\mathbf{p}^{1:N} \in \mathbb{R}^{N \times c}$, the extracted periodic motion features $\mathbf{f}^{1:N} \in \mathbb{R}^{N \times h}$. We choose $n=256$, $c=(6+3) \times 3 \times 2 = 54$ following previous works, and set the number of latent periodic channels $h$ to $6$. We set the input sequence length $N$ to $120$.

We train our VAE-based motion prior and conditional denoiser with a batch size of $64$ and use AdamW for tuning parameters. The learning rate is fixed to $0.001$ for both models. The feature dimension $d_{\text{model}}$ of our models are set to $256$, and the stacked transformer layers of the VAE-based motion prior and conditional denoiser are $9$ and $8$, respectively.

To keep the scale of the guidance score at the same level, we set the scaling factor $\lambda_{\text{penetration}}$ for scene-penetration loss $\ell_{\text{penetration}}$ to $0.1$, and the scaling factor $\lambda_{\text{phase}}$ for phase-matching loss $\ell_{\text{phase}}$ to $0.01$. We observe larger order of magnitude of scaling factors will result in performance degradation and severe jittering of generated motions.

\section{Extra qualitative results}

\begin{figure}
    \centering
    \resizebox{\columnwidth}{!}{\includegraphics{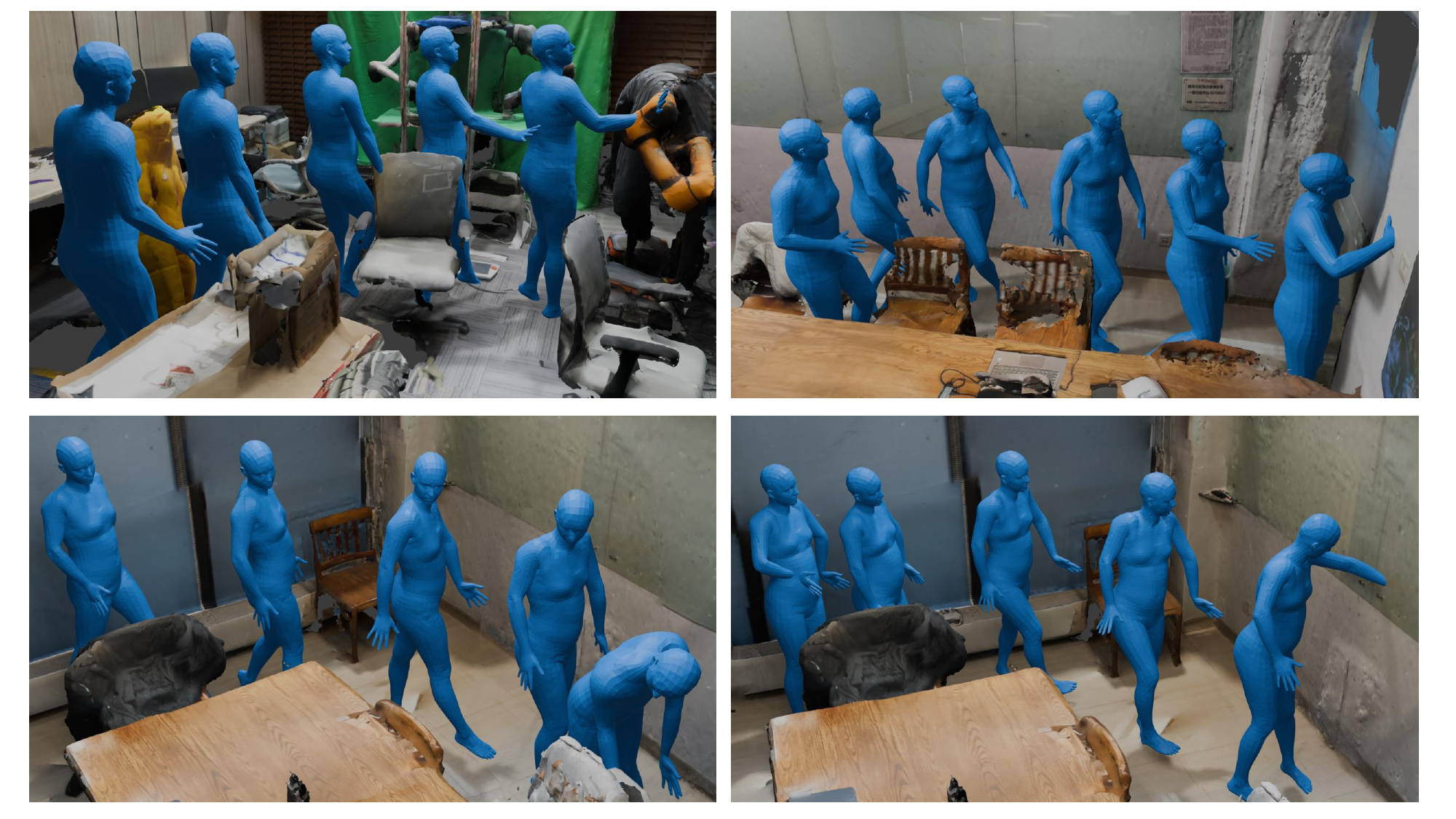}}
    \caption{The failure cases of our motion generation pipeline.}
    \label{fig:supp-failure}
\end{figure}

\begin{figure*}
    \centering
    \resizebox{2\columnwidth}{!}{\includegraphics{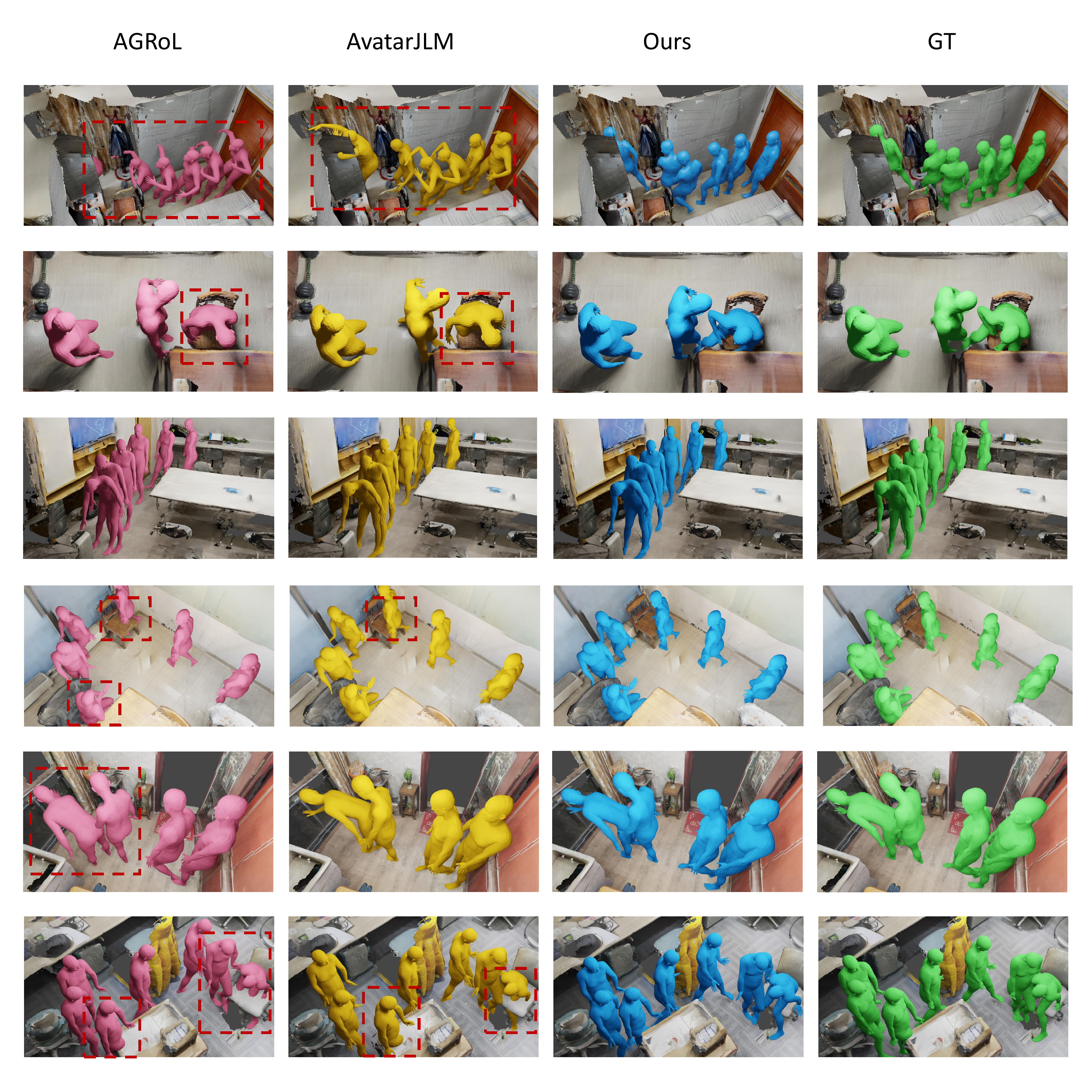}}
    \caption{The extra qualitative experiment evaluated on the GIMO dataset.}
    \label{fig:supp-exp}
\end{figure*}

We show the generated motions of our method against others on the GIMO dataset in Figure \ref{fig:supp-exp}. We highlight the implausible motions in rectangle marks, it is clear that our method learns the correct human-scene interactions and avoids scene penetration as much as possible. \\

\noindent \textbf{Failure cases and analysis.} We also show the failure cases of our motion generation pipeline in Figure \ref{fig:supp-failure}. While focusing on generating realistic lower body motions, our method failed to faithfully capture fine-grained hand-object interactions, such as picking up clothes or wiping the blackboard. Incorporating more sophisticated full-body physical constraints may resolve the failure cases and be considered in our future work.

\section{Extra evaluation of scene modality}
In this section, we evaluate the effect of the scene modality on the task of reconstructing full-body motion from \textit{head motion} only. The sparser inputs make the reconstruction task even more difficult. We compare our scene-conditioned diffusion backbone with a recent method EgoEgo which estimates full-body motion from egocentric videos. For a fair comparison, we provide ground truth head motions to EgoEgo and use its conditional diffusion network to generate full-body motions.

\input{tables/supp-main}

The quantitative results are shown in Table \ref{table:supp}, where we report the same metrics as EgoEgo. Although using similar diffusion backbones, by using extra scene modality, our method has higher estimation accuracy, showcasing the benefit of incorporating scene information.

\section{Human perceptual study}
We conducted a human perceptual study to investigate the quality of the motions generated by our model. We invite 25 users to provide three comparisons. For each comparison, we ask the users \textit{"Which of the two motions is more realistic?"}, and each user is provided 10 sequences to evaluate.

The results are shown in Figure \ref{fig:supp-perceptual-circle} and Figure \ref{fig:supp-perceptual-gimo}. Our results were preferred over the other state-of-the-art and are even competitive with ground truth motions on the CIRCLE dataset.

%% file: tables/supp-main.tex
\begin{table}
\centering
\resizebox{\columnwidth}{!}{\begin{tabular}{ccccccc}
\hline
& \multicolumn{3}{c}{GIMO \cite{zheng2022gimo}} & \multicolumn{3}{c}{CIRCLE \cite{Araujo_2023_CVPR}} \\ 
\hline
\bfseries Method & \textbf{MPJPE} & \textbf{Accel} & \textbf{FS} & \textbf{MPJPE} & \textbf{Accel} & \textbf{FS} \\
\hline
EgoEgo & 125.7 & 10.2 & 1.7 & 96.9 & 8.3 & 2.0 \\
Ours & \textbf{108.1} & \textbf{10.1} & 1.7 & \textbf{73.5} & \textbf{7.5} & \textbf{1.8} \\
\hline
\end{tabular} 
}
\caption{Full-body motion estimation results evaluated on GIMO \cite{zheng2022gimo} and CIRCLE \cite{Araujo_2023_CVPR}, given head motion only.}
\label{table:supp}
\end{table}